\documentclass[11pt]{article}

\usepackage[preprint]{acl}
\usepackage{times}
\usepackage{latexsym}
\usepackage[T1]{fontenc}
\usepackage[utf8]{inputenc}
\usepackage{microtype}
\usepackage{graphicx}
\usepackage{booktabs}
\usepackage{amsmath}
\usepackage{xcolor}
\usepackage{multirow}

\title{Configurable Clinical Information Extraction with Agentic RAG: What Works, What Breaks, and Why}

\author{
Osman Alperen \c{C}inar-Kora\c{s}$^{1,2}$\thanks{Corresponding author.} \quad
Marie Bauer$^{1}$ \quad
Sameh Khattab$^{1,2}$ \quad
Merlin Engelke$^{1}$ \quad
Moon Kim$^{1}$  \\
\bfseries Stephan Settelmeier$^{6}$ \quad
Shigeyasu Sugawara$^{1,5}$ \quad
Fabian Freisleben$^{1}$ \quad
Felix Nensa$^{1}$ \quad
Jens Kleesiek$^{1,2,3,4}$ \\[8pt]
$^{1}$Institute for Artificial Intelligence in Medicine (IKIM), University Medicine Essen, Essen, Germany \\[2pt]
$^{2}$Faculty of Computer Science, University of Duisburg-Essen, Essen, Germany \\[2pt]
$^{3}$Department of Physics, TU Dortmund University, Dortmund, Germany \\[2pt]
$^{4}$Lamarr Institute for Machine Learning and Artificial Intelligence, TU Dortmund University, Germany \\[2pt]
$^{5}$Advanced Clinical Research Center, Fukushima Medical University, Fukushima, Japan \\[2pt]
$^{6}$Department of Cardiology and Vascular Medicine, University Hospital Essen, Essen, Germany
}

\begin{document}
\maketitle

\begin{abstract}
Patient contexts span hundreds of heterogeneous documents and thousands of structured data points, yet the document-level metadata that AI systems need for retrieval and triage is absent or incomplete. Standard retrieval-augmented generation fails on this data, mishandling temporal reasoning, cross-document dependencies, and missing metadata. We deploy ACIE (Agentic Clinical Information Extraction) at University Medicine Essen: an on-premise agentic RAG pipeline that reasons over complete patient contexts and grounds every answer in source passages for clinician verification. We quantify the metadata gap, trace the architectural decisions it shaped, and evaluate extraction alongside an independent retrospective lymphoma registry study, in which nuclear-medicine physicians verify every extracted value against its cited sources. Across 7,326 judgments, clinicians accepted 96.5\% of extractions, with per-type acceptance ranging from 80\% to 99\%.
\end{abstract}

\section{Introduction}
\label{sec:intro}

Clinical workflows routinely require structured data compiled from patient records spanning thousands of documents and tens of thousands of structured data points across multiple hospital systems. Enrolling a single lymphoma patient in a clinical study, for example, requires reconstructing the treatment history and locating diagnostic markers across years of documents that may be duplicated, misdated, or buried among unrelated records. Clinicians perform this compilation by hand, and information is routinely missed \cite{moon2022}.

Clinical information extraction (IE) has long aimed to alleviate this burden, yet even recent deployed systems require developer effort to adapt to new workflows (\S\ref{sec:related}). Large language models can perform extraction without task-specific training \cite{singhal2023large, agrawal2022}, but LLM-based clinical IE remains largely confined to research evaluations \cite{artsi2025clinical}. Two barriers explain the gap. First, transmitting patient data to external servers raises privacy and regulatory risks prompting for on-premise deployment \cite{dennstaedt2025onpremise}. Second, real patient records pose retrieval challenges that standard RAG is not designed for, because the metadata it depends on is unreliable, documents are interdependent, and conflicting values require temporal reasoning to resolve. A recent scoping review found that only 9\% of end-to-end medical RAG systems employ agentic architectures \cite{miao2025}.

We deploy ACIE (Agentic Clinical Information Extraction) at University Medicine Essen, whose FHIR repository, with nearly 2 billion resources, is among the largest in Europe. Clinicians define extraction schemas with typed targets without developer involvement. An agentic RAG pipeline reasons over complete patient contexts, grounding every value in source passages for clinician verification, running entirely on-premise. Our contributions are:

\textbf{1.} A clinician-verified evaluation of agentic extraction alongside an independent retrospective lymphoma registry study (74 clinician-configured fields, 99 patients, 7,326 judgments), in which nuclear-medicine physicians accept or reject every extracted value and label rejections with structured error and editorial categories.

\textbf{2.} A quantified analysis of the metadata gap between what AI systems need and what clinical data exports provide.

\textbf{3.} Architectural decisions shaped by this data reality, illustrating the design trade-offs of building on real clinical data.

\section{Related Work}
\label{sec:related}

\textbf{From rules to domain-specific pretraining.}
Early clinical IE relied on engineered NLP pipelines such as cTAKES \cite{savova2010}, where extraction targets were defined by developers. Knowledge Author \cite{scuba2016} enabled domain experts to define schemas through a web interface, but its rule-based backend limited expressiveness: only 76\% of target concepts could be fully represented, with recall as low as 46\%. Domain-specific pretraining (BioBERT \cite{lee2020}, GatorTron \cite{yang2022}) improved accuracy but still required fine-tuning per extraction target. In a survey of 263 clinical IE studies, \cite{wang2018clinical} found that over half targeted disease-related extraction spanning 88 unique diseases, concluding that the portability and generalizability of clinical IE systems are still limited. Community shared tasks from i2b2/VA \cite{uzuner2011} and n2c2 \cite{henry2019} to recent iterations \cite{lybarger2023, yao2024chemo} similarly operate on predefined targets. Throughout this, extraction targets remained fixed or configurability mechanisms could not meet the demands of real clinical complexity.

\textbf{LLMs shift what is possible but remain largely undeployed.}
LLMs enabled few-shot clinical extraction without task-specific training, yet LLM-based clinical IE has rarely moved beyond research evaluations \cite{artsi2025clinical}.
\cite{wiest2024} locally deploy Llama~2 for five fixed features from MIMIC-IV patient histories, demonstrating on-premise feasibility but with static, researcher-defined targets.
LLM-AIx \cite{wiest2025llmaix} provides an open-source pipeline for structured extraction from individual documents with user-defined schemas and local inference, but processes documents independently without retrieval augmentation and has only been validated on research datasets.
Deployed systems have followed a separate track. MedCAT \cite{kraljevic2021} and MiADE \cite{jiangkells2025} are clinical NLP pipelines in production, but with developer-defined targets. Griot \cite{griot2025} deploys Qwen3-235B with RAG inside Epic for clinical assistance (1,028 users) and Gr{\"u}nig et al.~\cite{gruenig2025} deploy an on-premise LLM at a German university hospital, but neither performs structured extraction.

\textbf{Agentic RAG as the emerging frontier.}
Retrieval-augmented generation \cite{lewis2020} grounds LLM outputs in retrieved evidence, and agentic frameworks like ReAct \cite{yao2023} enable iterative reasoning over complex information needs.
i-MedRAG \cite{xiong2024} shows that iterative retrieval outperforms single-pass RAG for medical QA but uses a fixed iteration schedule on curated knowledge bases.
Agentic clinical IE has recently emerged: CLINES \cite{clines2025} structures clinical concepts through a modular pipeline, HARMON-E \cite{harmone2025} applies hierarchical reasoning to oncology notes, and ReflecTool \cite{reflectool2025} benchmarks tool-augmented clinical agents. However, all use researcher-defined targets on benchmark datasets; none are deployed with clinician-configurable schemas.
To our knowledge, no prior work has quantified the gap between the metadata AI systems need and what clinical data exports provide, or traced architectural decisions to data quality failures in a deployed system.

\section{System Overview}
\label{sec:system}

\begin{figure*}
\centering
\includegraphics[width=\textwidth]{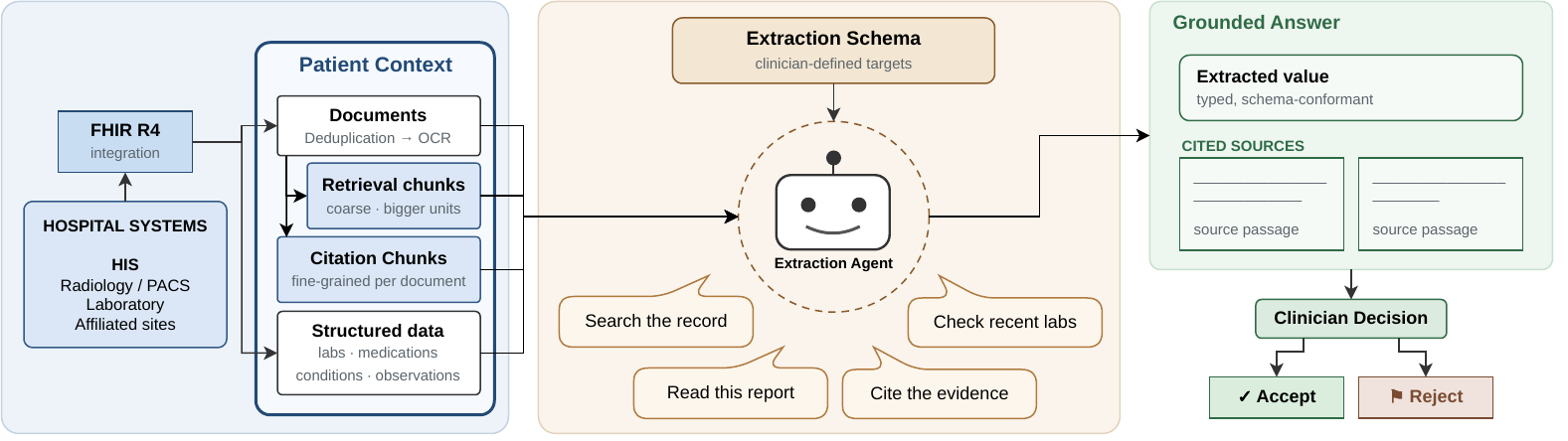}
\caption{ACIE system overview. Clinical data from multiple hospital systems, accessed via a FHIR server, is organized into a patient context. Each document is chunked at two granularities for retrieval and citation. For each extraction target, an agent iteratively searches and inspects the patient context, returning a grounded answer that the clinician verifies against cited source passages. The structured-data labels indicate FHIR resource types.}
\label{fig:system}
\end{figure*}

Figure~\ref{fig:system} illustrates the pipeline. Clinicians configure extraction targets through typed schemas, and the system handles retrieval and extraction. This section describes the deployed system. \S\ref{sec:lessons} traces the architectural decisions behind it. Implementation details of the extraction schema engine, agent orchestration, and document export pipeline are outside the scope of this paper.

\subsection{Patient Context}
\label{sec:retrieval}

A patient context consists of all clinical data available for a patient: documents (discharge letters, radiology reports, laboratory findings, referral letters, operative notes) and structured FHIR \cite{hl7fhir} data points (laboratory results, medications, conditions, observations) accumulated over years of care. \S\ref{sec:dataquality} characterizes the scale and quality of this data.

ACIE ingests all available data via the hospital's FHIR server. Non-machine-readable documents are processed via OCR, and those falling below a quality threshold are excluded. Each document is semantically chunked at two granularities: coarse \emph{retrieval chunks} that preserve surrounding context, and fine-grained passages serving as atomic units for source citation. Clinical documents interleave narrative, tabular, and form-like content in heterogeneous layouts. Chunking their serialized text yields many short, low-information fragments that dense retrievers systematically favor \cite{fayyaz2025}. We apply a length-penalized retrieval score:
\begin{equation}
\label{eq:rerank}
s = \text{sim}(q, c) \cdot p(\ell), \quad p(\ell) = \min\!\left(\frac{\ell}{\tau},\; 1\right) \cdot \tfrac{2}{3} + \tfrac{1}{3}
\end{equation}
where $\text{sim}(q, c)$ is the cosine similarity between the query $q$ and chunk $c$, $\ell$ is the chunk length in characters, and $\tau = 40$. The penalty $p(\ell)$ dampens scores of short fragments without eliminating them. We fixed $\tau$ and the blend weights on a development subset and held them constant across tasks.

\subsection{Agentic Extraction}
\label{sec:agentic}

Clinicians define extraction targets through a typed schema. The same system serves different clinical use cases (pre-procedure protocols, retrospective study data collection, clinical documentation) through schema configuration alone, without code changes.

For each extraction target, a tool-calling agent \cite{yao2023} searches the patient context. Standard retrieve-then-generate pipelines with metadata-based filters proved insufficient because the metadata they depend on is unreliable or absent (\S\ref{sec:dataquality}, \S\ref{sec:lessons}). The agent's tools allow searching by semantic similarity across the full patient context, listing documents with query-relevant summaries, inspecting a specific document in detail, and querying structured data directly. When listing documents, summaries are generated on the fly by assembling the highest-scoring citation chunks per document in document order until at least 200 words are accumulated, producing a chronologically coherent, content-based relevance preview.

The agent iterates until it has gathered sufficient evidence, then returns an answer with every value attributed to specific source passages. This grounding is a safety requirement: clinicians review each extracted value against the cited passages and accept or reject it before it enters clinical documentation.

\subsection{Deployment}
\label{sec:deployment}

ACIE runs entirely on-premise on hospital infrastructure, deployed as a web application on Kubernetes. Patient data never leaves the hospital network. The extraction model is Qwen~3.6 35B-A3B \cite{qwen3.6}, a mixture-of-experts model. Scanned documents are processed by PaddleOCR-VL~1.5 \cite{paddleocr}. For this evaluation, the extraction model was served on 4$\times$H100 GPUs and the OCR model on a single H100 GPU.

\section{Evaluation}
\label{sec:eval}

ACIE is deployed at University Medicine Essen, whose FHIR~R4 server, conforming to a national interoperability core-dataset specification, integrates nearly 2 billion resources across 1.7 million patients from the hospital's primary information system, radiology, laboratory, and affiliated hospitals, making it one of the largest clinical FHIR repositories in Europe. Despite this scale, document-level metadata remains sparse (\S\ref{sec:dataquality}). Table~\ref{tab:corpus} summarizes the corpus, Table~\ref{tab:patients} characterizes patient contexts from 10,000 randomly sampled patients.

\subsection{FHIR Data Quality Analysis}
\label{sec:dataquality}

We characterize the challenges clinical data poses for automated extraction across 10,000 patients ($\sim$1.2M deduplicated documents).

\textbf{Encounter linkage and distribution.}
FHIR groups clinical activities into encounters, which could in principle organize a patient's documents by episode of care. In this export, encounters follow a three-level hierarchy (case, department, stay), but documents link exclusively to case-level encounters, the broadest administrative unit. Of these, 13.7\% hold no documents at all, and the remainder are highly non-uniform: a single encounter holds a median of 47.5\% of a patient's documents (Table~\ref{tab:concentration}), dropping to 14.7\% for patients with 20+ encounters, which suggests the hierarchy distributes documents as intended (Appendix~\ref{sec:appendix-enc-coverage}). Yet at P99, coverage remains 53.5\% even for patients with 20+ encounters, and the concentration index reaches 14.83 (Table~\ref{tab:concentration}): the most complex patients are precisely those where encounter structure fails to partition documents meaningfully. Linkage is also temporally imprecise: 56.5\% of linked documents carry timestamps entirely outside their encounter's period (median delta 14.0 days). Heuristics may partially recover episode-level structure, but cannot guarantee reliable scoping, which is why we bypass encounter-based scoping altogether (\S\ref{sec:lessons}).

\begin{table}
\centering
\small
\begin{tabular}{lr}
\toprule
\textbf{Category} & \textbf{Count} \\
\midrule
Patient records & 5,598,272 \\
\quad Unique individuals & 1,747,135 \\
\midrule
Orders and requests & 852M \\
Lab values and observations & 675M \\
Clinical documents & 84M \\
Medication orders & 77M \\
Diagnoses and conditions & 40M \\
Medication administrations & 33M \\
Clinical encounters & 29M \\
Other & 182M \\
\midrule
\textbf{Total FHIR resources} & \textbf{1.97B} \\
\bottomrule
\end{tabular}
\caption{Clinical data corpus. Only the most populated resource types are shown.}
\label{tab:corpus}
\end{table}

\begin{table}
\centering
\small
\begin{tabular}{lrrr}
\toprule
& \textbf{Med.\ (IQR)} & \textbf{P99} & \textbf{Max} \\
\midrule
Docs & 52 (14--140) & 937 & 2,542 \\
Dedup.\ (\%) & 33.5 (20.0--43.0) & -- & 54.6 \\
Struct.\ resources & 406 (38--1,922) & 37,074 & 119,191 \\
Encounters & 18 (6--46) & 323 & 1,207 \\
OCR rej.\ (\%) & 10.3 (6.8--17.8) & -- & 52.0 \\
\bottomrule
\end{tabular}
\caption{Per-patient statistics ($n$=10,000), sampled randomly from 2025. Docs = deduplicated documents; Dedup.\ = fraction of raw documents removed by deduplication; Struct.\ resources = non-document FHIR resources (lab values, medications, conditions, etc.); OCR rej.\ = fraction of documents rejected by OCR quality filtering.}
\label{tab:patients}
\end{table}

\textbf{Document quality and metadata.}
FHIR provides mechanisms for document relationships and unique identifiers but only recommends them: just 0.52\% and 27.8\% of documents carry them. Content-level deduplication removes a median 33.5\% of documents per patient (up to 54.6\%; Table~\ref{tab:patients}). Metadata on the FHIR document reference is otherwise sparse: authorship appears for 1.9\%, subtypes for 41.87\%, structured conclusions for 0.45\%. Provenance fields are better populated on other resource types (e.g., 97.5\% on diagnostic reports; Appendix, Table~\ref{tab:metadata-population}), but these do not cover the full document corpus. Crucially, no document-level summary or abstract exists that a retrieval or agentic pipeline could use as a content preview to decide whether a document is worth reading. Over 1,000 document categories are used, many differing only in wording, and OCR rejection reaches 52.0\% for the worst patient (median 10.3\%; Table~\ref{tab:patients}).

\textbf{Timestamp reliability.}
Documents carry several metadata timestamps (report finalization, file creation, record update, encounter). The most clinically meaningful, the encounter timeframe, is absent from the export, and not propagated to the document reference. Even when inferred from the linked encounter resource, 56.5\% of documents carry timestamps outside their encounter period. We therefore resolve each document's primary date from the available timestamps via a priority cascade. To assess whether the resolved timestamp reflects the actual clinical date, we compared it against the date extracted from document content via OCR and an LLM. When no date can be identified from the content, the system falls back to the resolved timestamp, so reported agreement is an upper bound.
Only 58.8\% agreed on the same day, and 36.5\% diverged by more than one day. Agreement stays near 59\% whichever field supplies the date (Appendix~\ref{sec:appendix-crossval}), so no document-level timestamp reliably represents the clinical date or orders a patient's context in time.

\textbf{Patient context scaling.}
Patient contexts span orders of magnitude: after deduplication, document counts range from 1 to over 2,500, structured data points from 0 to over 119,000, and document lengths from 24 to over 900,000 characters (Appendix~\ref{sec:appendix-perpatient}, Table~\ref{tab:perpatient-full}). The top 1\% (P99) define the hardest cases any deployed system must handle without degradation. These patient contexts hold at least 937 documents and over 37,000 structured resources. Appendix~\ref{sec:appendix-history} provides a breakdown by history length.

\subsection{Clinical Study Extraction}
\label{sec:study-eval}

We evaluate ACIE alongside an independent retrospective registry study of lymphoma patients undergoing molecular imaging (Appendix~\ref{sec:appendix-study}). Its electronic case report form (eCRF), designed by two nuclear-medicine physicians and a hematologist, predates ACIE and was defined independently of it, so the extraction targets were not shaped by what the tool can do. A nuclear-medicine specialist with over four years of training configured all 74 AI-extracted fields (45 categorical, 9 numerical, 8 Boolean, 6 date, 3 free-text, 3 tabular; Appendix~\ref{sec:appendix-study}), choosing their typed specifications and refining them on a subset of cohort patients, with no engineering effort. The fields cover clinical classification, immunohistochemical and molecular markers, longitudinal treatment and imaging history, and outcomes.

ACIE extracted these fields for 99 patients (7,326 values). Each value and its cited passages were verified by one of two nuclear-medicine physicians against the clinical systems used in routine work, so acceptance estimates verified correctness rather than agreement with a fixed key. Rejections are typed as \emph{extraction errors} (wrong, fabricated, extraneous, or missed), \emph{editorial adjustments} (acceptable value, but more or less detail wanted), or \emph{form configuration} issues (Table~\ref{tab:rejections}). We report pooled rates alongside the patient-level distribution.

\textbf{Reliability.}
The blended acceptance rate (7,073 of 7,326, 96.5\%) combines two behaviours. Where the system committed to a value (4,440 fields) clinician-verified precision is 96.4\%; where it returned nothing (2,886 fields) 96.8\% were correct abstentions, leaving 92 in which a value did in fact exist (per type, Appendix~\ref{sec:appendix-erroranalysis}, Table~\ref{tab:abstention}). It thus neither produces extraneous values for fields that should be empty, a common LLM failure, nor misses present ones in bulk, and is consistent across patients (mean 96.5\%, median 97.3\%, range 82.4--100\%; 78 of 99 patients $\geq$95\%, 7 patients with no rejections).

\textbf{Field type drives accuracy.}
Acceptance varies far more by data type than the 74-field count suggests (Table~\ref{tab:acctype}). The categorical, numerical, Boolean, and free-text fields are accepted at 96.0--98.6\%, whereas the two weak types are dates at 84.3\% and tabular fields at 79.8\% (71.2\% among non-empty tables), and the rejected cases show why. Tabular fields are rejected even when their evidence is directly stated in the sources: the difficulty is assembling the multi-row tabular timeline itself, with missing or guessed dates, dropped rows, and extraneous ones. Dates fail in two directions, when a value is returned the reviewer often judges it the wrong clinical event among several candidates, and when none is returned a value frequently did exist, making date the one type whose abstentions are unreliable (empty answers accepted only 69.8\%, against 96.8\% across types). The two hardest individual fields are of exactly these kinds, the date of death or last follow-up (accepted for 55 of 99 patients) and the treatment-timeline table (59 of 99). The driver of error is thus how much temporal reasoning the answer demands over the full record, assembling a timeline for tables or selecting the right event for dates.

\begin{table}
\centering
\small
\begin{tabular}{lrrr}
\toprule
\textbf{Field type} & \textbf{$n$} & \textbf{Acc.\%} & \textbf{Empty\%} \\
\midrule
Categorical & 4,455 & 98.6 & 35.1 \\
Numerical & 891 & 98.3 & 76.1 \\
Boolean & 792 & 98.6 & 21.8 \\
Free text & 297 & 96.0 & 59.9 \\
Date & 594 & 84.3 & 34.0 \\
Tabular & 297 & 79.8 & 31.0 \\
\midrule
\textbf{Total} & \textbf{7,326} & \textbf{96.5} & \textbf{39.4} \\
\bottomrule
\end{tabular}
\caption{Clinician acceptance by field type. ``Empty\%'' is the share of fields where the system returned no value, typically a legitimately absent value.}
\label{tab:acctype}
\end{table}

\textbf{Errors and safety.}
Of 253 rejections (Table~\ref{tab:rejections}), 241 are extraction errors, 3 are editorial adjustments, and 9 are form configuration issues (e.g., dropdown options not matching clinical reality) rather than extraction failures; excluding the latter, extraction-attributable acceptance is 96.7\%. By direction of failure, 161 rejections correct a value the system produced and 92 supply one it left empty (overwhelmingly dates). The errors are also highly concentrated: ten of the 74 fields account for 182 of the 253 rejections (Appendix~\ref{sec:appendix-erroranalysis}), led by the death-or-last-follow-up date and the treatment timeline. No value was hallucinated, i.e.\ produced without support from any cited passage. In a single case the system returned a value where it should have abstained (one extraneous extraction). The dominant residual risk is thus a wrong or missing value that a reviewer corrects, not an invented one.

\begin{table}
\centering
\small
\begin{tabular}{lr}
\toprule
\textbf{Rejection category} & \textbf{Count (\%)} \\
\midrule
\multicolumn{2}{l}{\emph{Extraction errors}} \\
\quad Incorrect value (needs correction) & 221 (87.4) \\
\quad Fully incorrect (replaced) & 7 (2.8) \\
\quad Missed extraction (false negative) & 9 (3.6) \\
\quad Missing reference & 3 (1.2) \\
\quad Extraneous extraction (false positive) & 1 (0.4) \\
\quad Hallucinated & 0 (0.0) \\
\midrule
\multicolumn{2}{l}{\emph{Editorial adjustments}} \\
\quad Missing information & 2 (0.8) \\
\quad Excess information & 1 (0.4) \\
\midrule
\multicolumn{2}{l}{\emph{Form configuration}} \\
\quad Configuration error & 9 (3.6) \\
\bottomrule
\end{tabular}
\caption{The 253 rejected fields by category (\% of all rejections); category definitions in Appendix~\ref{sec:appendix-rejcat}. Each rejected field carried exactly one flag. Extraction errors dominate; no content was hallucinated.}
\label{tab:rejections}
\end{table}

\section{Lessons from Deployment}
\label{sec:lessons}

We report two lessons from deploying ACIE on real clinical data, quantifying the specific mechanisms we encountered.

\textbf{L1: Clinical data quality falls far short of what AI systems require.}

Prior work documents clinical data quality challenges \cite{vorisek2022} and heterogeneity across institutions \cite{palm2025}. \S\ref{sec:dataquality} quantifies this gap from the perspective of an AI system that must retrieve and extract based on this metadata. The deployment site operates a large-scale clinical FHIR repository, yet document-level metadata remains sparse: fields like encounter periods, document relationships, and authorship are absent from the document reference or too sparsely populated for filtering. Clinicians navigate this sparsity through institutional knowledge, but AI systems cannot. In a large primary care database, only 13\% of clinical concepts in free-text notes had structured counterparts in coded fields \cite{seinen2024}, and where timestamps are populated, they are not necessarily correct (\S\ref{sec:dataquality}). These gaps are likely not site-specific: independent single-site studies document data quality problems of similar severity in other settings, from pervasive duplication in US clinical notes \cite{steinkamp2022} to decades of erroneous administrative entries at a regional German hospital \cite{forstel2024}. We hypothesize that this reflects data infrastructure historically optimized for billing rather than clinical coherence. Any system deployed on such data must compensate architecturally.

\textbf{L2: Architectural decisions shaped by data.}

The data quality gaps in \S\ref{sec:dataquality} shaped three design decisions in ACIE's architecture.

\textbf{Agentic retrieval over static filtering.}
We first deployed a retrieve-then-generate pipeline with metadata-based filters (encounter-scoped retrieval, date-range and category filters). Each filter depended on metadata that \S\ref{sec:dataquality} shows is unreliable or absent. After exhausting static filter combinations, we concluded that reliable retrieval requires an agent that reasons about which documents matter based on content, not a pipeline that filters by metadata.

\textbf{Query-relevant document summaries.}
To make content-based triage tractable over hundreds of documents, the agent cannot read every document in full. Instead, it previews documents through the query-relevant summaries of \S\ref{sec:agentic} and decides from content whether full inspection is warranted. Fine-grained chunking compounds the problem by producing fragments that vary by orders of magnitude in length, which dense retrievers systematically favor when short \cite{fayyaz2025}. The length penalty in Equation~\ref{eq:rerank} corrects this bias.

\textbf{Markdown over JSON serialization.}
Data serialization affects extraction quality for smaller models \cite{pator2026}, and output format constraints degrade reasoning \cite{tam2024}. We encountered a related failure on the \emph{input} side: when patient metadata was presented in JSON format, specific patients consistently triggered malformed tool calls. The failures were deterministic and patient-specific, suggesting that particular JSON structures from heterogeneous clinical metadata interfered with tool-calling. Switching to markdown eliminated all failures.

\section{Conclusion}
\label{sec:conclusion}
Clinical IE research typically treats the data as given and asks how capable the model is. Deploying ACIE taught us the inverse: the data dictates the architecture. The document-level metadata needed for retrieval and triage is largely absent, unreliable or not propagated to the document level even at a site with large-scale FHIR integration, so retrieval cannot filter by structure, and reasoning must move into the retrieval loop. Evaluated alongside an independent lymphoma registry study whose extraction targets predate the system, ACIE reached 96.5\% acceptance with no hallucinated content. The residual errors were governed by the temporal reasoning each target demanded. Grounding every value in source passages shifts the clinician's role from compiling to verifying. Extraction can run in batches outside working hours, and the study physicians reported roughly three times faster completion per patient. These metadata gaps are well-documented across institutions \cite{steinkamp2022, forstel2024}, and we expect any system deployed on real hospital data to face similar constraints. Deployed clinical extraction therefore rests on two supports: architectures that reason over content, and human verification of grounded outputs.

\section*{Limitations}

Our evaluation is a single retrospective study, one disease area, one hospital, one language, and 99 patients, so generalization to other settings is untested, and each field was graded by a single expert reviewer, leaving inter-rater reliability unmeasured. The retrospective setting is more permissive than point-of-care use, where a value directly informs an intervention rather than populating a research cohort; the headline acceptance also reflects the large share of fields whose correct answer is legitimately absent (39.4\%), with precision on produced values at 96.4\% and lower on the weakest types (84.3\% dates, 79.8\% tables). The absence of flagged hallucinations was judged by the same reviewers under source-grounded review rather than by independent adjudication of every passage. We did not compare against a non-agentic or commercial baseline, so we do not isolate the contribution of the agentic design, and the data-quality findings (\S\ref{sec:dataquality}) come from a single hospital's FHIR repository, so the specific rates may differ elsewhere. Extraction quality also remains bounded by the on-premise model.

\section*{Ethical Considerations}
ACIE runs entirely on-premise, so patient data never leaves the hospital network. It is assistive, not autonomous: every value is grounded in cited passages and must be verified by a clinician before it enters documentation. This mitigates but does not remove the risk of automation bias, where a frictionless interface invites uncritical acceptance; mandatory source review, and the rejection workflow we evaluate, are the safeguard. Extraction quality may degrade for patient groups underrepresented in the records or in the underlying model, a risk clinician verification is intended to catch. The registry study was conducted in accordance with the applicable institutional and regulatory requirements for the retrospective use of clinical data at University Medicine Essen.

In line with the ACL policy on AI writing assistance, an AI assistant was used for language editing and LaTeX formatting only; all research content and claims originate from the authors, who take full responsibility for the final text.
\bibliography{references}

@misc{qwen3.6,
  title={{Qwen3.6-35B-A3B}: Agentic Coding Power, Now Open to All},
  author={{Qwen Team}},
  month={April},
  year={2026},
  url={https://qwen.ai/blog?id=qwen3.6-35b-a3b},
  note={Model card: \url{https://huggingface.co/Qwen/Qwen3.6-35B-A3B}. Accessed: 2026-06-11.}
}

@article{paddleocr,
  title={{PaddleOCR-VL-1.5}: Towards a Multi-Task 0.9B {VLM} for Robust In-the-Wild Document Parsing},
  author={Cui, Cheng and Sun, Ting and Liang, Suyin and Gao, Tingquan and Zhang, Zelun and Liu, Jiaxuan and Wang, Xueqing and Zhou, Changda and Liu, Hongen and Lin, Manhui and Zhang, Yue and Zhang, Yubo and Liu, Yi and Yu, Dianhai and Ma, Yanjun},
  journal={arXiv preprint arXiv:2601.21957},
  year={2026}
}

@article{seinen2024,
  title={Using Structured Codes and Free-Text Notes to Measure Information Complementarity in Electronic Health Records: {Feasibility} and Validation Study},
  author={Seinen, Tom M. and Kors, Jan A. and van Mulligen, Erik M. and Rijnbeek, Peter R.},
  journal={Journal of Medical Internet Research},
  volume={27},
  pages={e66910},
  year={2025},
  doi={10.2196/66910}
}

@article{steinkamp2022,
  title={Prevalence and Sources of Duplicate Information in the Electronic Medical Record},
  author={Steinkamp, Jackson and Kantrowitz, Jacob J. and Airan-Javia, Subha},
  journal={JAMA Network Open},
  volume={5},
  number={9},
  pages={e2233348},
  year={2022},
  doi={10.1001/jamanetworkopen.2022.33348}
}

@article{forstel2024,
  title={Data quality in hospital information systems: {Lessons} learned from analyzing 30 years of patient data in a regional {German} hospital},
  author={F{\"o}rstel, Stefan and F{\"o}rstel, Markus and Gallistl, Markus and Zanca, Dario and Eskofier, Bjoern M. and Rothgang, Eva M.},
  journal={International Journal of Medical Informatics},
  volume={192},
  pages={105636},
  year={2024},
  doi={10.1016/j.ijmedinf.2024.105636}
}

@article{wang2018clinical,
  title={Clinical Information Extraction Applications: A Literature Review},
  author={Wang, Yanshan and Wang, Liwei and Rastegar-Mojarad, Majid and Moon, Sungrim and Shen, Feichen and Afzal, Naveed and Liu, Sijia and Zeng, Yuqun and Mehrabi, Saeed and Sohn, Sunghwan and Liu, Hongfang},
  journal={Journal of Biomedical Informatics},
  volume={77},
  pages={34--49},
  year={2018},
  publisher={Elsevier},
  doi={10.1016/j.jbi.2017.11.011}
}

@article{uzuner2011,
  title={2010 {i2b2/VA} Challenge on Concepts, Assertions, and Relations in Clinical Text},
  author={Uzuner, {\"O}zlem and South, Brett R. and Shen, Shuying and DuVall, Scott L.},
  journal={Journal of the American Medical Informatics Association},
  volume={18},
  number={5},
  pages={552--556},
  year={2011},
  doi={10.1136/amiajnl-2011-000203}
}

@article{henry2019,
  title={2018 {n2c2} Shared Task on Adverse Drug Events and Medication Extraction in Electronic Health Records},
  author={Henry, Sam and Buchan, Kevin and Filannino, Michele and Stubbs, Amber and Uzuner, {\"O}zlem},
  journal={Journal of the American Medical Informatics Association},
  volume={27},
  number={1},
  pages={3--12},
  year={2020},
  doi={10.1093/jamia/ocz166}
}

@article{lybarger2023,
  title={The 2022 {n2c2/UW} Shared Task on Extracting Social Determinants of Health},
  author={Lybarger, Kevin and Yetisgen, Meliha and Uzuner, {\"O}zlem},
  journal={Journal of the American Medical Informatics Association},
  volume={30},
  number={8},
  pages={1367--1378},
  year={2023},
  doi={10.1093/jamia/ocad012}
}

@inproceedings{yao2024chemo,
  title={Overview of the 2024 Shared Task on Chemotherapy Treatment Timeline Extraction},
  author={Yao, Liang and Hochheiser, Harry and Yoon, Wonjin and Goldner, Shyam and Savova, Guergana},
  booktitle={Proceedings of the 6th Clinical NLP Workshop},
  year={2024},
  doi={10.18653/v1/2024.clinicalnlp-1.53}
}

@article{kraljevic2021,
  title={Multi-domain Clinical Natural Language Processing with {MedCAT}: The Medical Concept Annotation Toolkit},
  author={Kraljevi{\'c}, Zeljko and Searle, Thomas and Shek, Anthony and Roguski, Lukasz and Noor, Kawsar and Bean, Daniel and Mascio, Aurelie and Zhu, Leilei and Folarin, Amos A. and Roberts, Angus and Bendayan, Rebecca and Richardson, Mark P. and Stewart, Robert and Shah, Anoop D. and Wong, Wai Keong and Ibrahim, Zina and Teo, James T. and Dobson, Richard J.B.},
  journal={Artificial Intelligence in Medicine},
  volume={117},
  pages={102083},
  year={2021},
  doi={10.1016/j.artmed.2021.102083}
}

@article{jiangkells2025,
  title={Design and implementation of a natural language processing system at the point of care: {MiADE} (medical information {AI} data extractor)},
  author={Jiang-Kells, Jennifer and Brandreth, James and Zhu, Leilei and Ross, Jack and Jani, Yogini and Costanza, Enrico and Amran, Maisarah and Kraljevi{\'c}, Zeljko and Bai, Xi and Dilan, M.M.N.S. and Wijayarathne, Jayathri and Wickramaratne, Ravi and Asselbergs, Folkert W. and Dobson, Richard J.B. and Wong, Wai Keong and Shah, Anoop D.},
  journal={BMC Medical Informatics and Decision Making},
  volume={25},
  number={1},
  pages={365},
  year={2025},
  doi={10.1186/s12911-025-03195-1}
}

@article{singhal2023large,
  title={Large Language Models Encode Clinical Knowledge},
  author={Singhal, Karan and Azizi, Shekoofeh and Tu, Tao and Mahdavi, S. Sara and Wei, Jason and Chung, Hyung Won and Scales, Nathan and Tanwani, Ajay and Cole-Lewis, Heather and Pfohl, Stephen and Payne, Perry and Seneviratne, Martin and Gamber, Paul and Kelly, Chris and Babiker, Abubakr and Sch{\"a}rli, Nathanael and Chowdhery, Aakanksha and Mansfield, Philip and Aguera y Arcas, Blaise and Webster, Dale and Corrado, Greg S. and Matias, Yossi and Chou, Katherine and Gottweis, Juraj and Tomasev, Nenad and Liu, Yun and Rajkomar, Alvin and Barral, Joelle and Semturs, Christopher and Karthikesalingam, Alan and Natarajan, Vivek},
  journal={Nature},
  volume={620},
  number={7972},
  pages={172--180},
  year={2023},
  doi={10.1038/s41586-023-06291-2}
}

@inproceedings{agrawal2022,
  title={Large Language Models are Few-Shot Clinical Information Extractors},
  author={Agrawal, Monica and Hegselmann, Stefan and Lang, Hunter and Kim, Yoon and Sontag, David},
  booktitle={Proceedings of the 2022 Conference on Empirical Methods in Natural Language Processing},
  pages={1998--2022},
  year={2022},
  publisher={Association for Computational Linguistics},
  doi={10.18653/v1/2022.emnlp-main.130},
  url={https://aclanthology.org/2022.emnlp-main.130}
}

@article{artsi2025clinical,
  title={Large Language Models in Real-World Clinical Workflows: A Systematic Review of Applications and Implementation},
  author={Artsi, Yaara and Sorin, Vera and Glicksberg, Benjamin S. and Korfiatis, Panagiotis and Nadkarni, Girish N. and Klang, Eyal},
  journal={Frontiers in Digital Health},
  volume={7},
  pages={1659134},
  year={2025},
  doi={10.3389/fdgth.2025.1659134}
}

@article{dennstaedt2025onpremise,
  title={Implementing Large Language Models in Healthcare While Balancing Control, Collaboration, Costs and Security},
  author={Dennst{\"a}dt, Fabio and Hastings, Janna and Putora, Paul Martin and Schmerder, Max and Cihoric, Nikola},
  journal={npj Digital Medicine},
  volume={8},
  number={1},
  pages={143},
  year={2025},
  doi={10.1038/s41746-025-01476-7}
}

@article{miao2025,
  title={Improving Large Language Model Applications in the Medical and Nursing Domains With Retrieval-Augmented Generation: Scoping Review},
  author={Miao, Yiqun and Zhao, Yuhan and Luo, Yuan and Wang, Huiying and Wu, Ying},
  journal={Journal of Medical Internet Research},
  volume={27},
  number={1},
  pages={e80557},
  year={2025},
  doi={10.2196/80557}
}

@article{savova2010,
  title={Mayo Clinical Text Analysis and Knowledge Extraction System ({cTAKES}): Architecture, Component Evaluation and Applications},
  author={Savova, Guergana K. and Masanz, James J. and Ogren, Philip V. and Zheng, Jiaping and Sohn, Sunghwan and Kipper-Schuler, Karin C. and Chute, Christopher G.},
  journal={Journal of the American Medical Informatics Association},
  volume={17},
  number={5},
  pages={507--513},
  year={2010},
  doi={10.1136/jamia.2009.001560}
}

@article{scuba2016,
  title={Knowledge Author: Facilitating User-driven, Domain Content Development to Support Clinical Information Extraction},
  author={Scuba, William and Tharp, Melissa and Mowery, Danielle and Tseytlin, Eugene and Liu, Yang and Drews, Frank A. and Chapman, Wendy W.},
  journal={Journal of Biomedical Semantics},
  volume={7},
  pages={42},
  year={2016},
  doi={10.1186/s13326-016-0086-9}
}

@article{lee2020,
  title={{BioBERT}: A Pre-trained Biomedical Language Representation Model for Biomedical Text Mining},
  author={Lee, Jinhyuk and Yoon, Wonjin and Kim, Sungdong and Kim, Donghyeon and Kim, Sunkyu and So, Chan Ho and Kang, Jaewoo},
  journal={Bioinformatics},
  volume={36},
  number={4},
  pages={1234--1240},
  year={2020},
  doi={10.1093/bioinformatics/btz682}
}

@article{yang2022,
  title={A Large Language Model for Electronic Health Records},
  author={Yang, Xi and Chen, Aokun and PourNejatian, Nima and Shin, Hoo Chang and Smith, Kaleb E. and Parisien, Christopher and Compas, Colin and Martin, Cheryl and Costa, Anthony B. and Flores, Mona G. and Zhang, Ying and Magoc, Tanja and Harle, Christopher A. and Lipori, Gloria and Mitchell, Duane A. and Hogan, William R. and Shenkman, Elizabeth A. and Bian, Jiang and Wu, Yonghui},
  journal={npj Digital Medicine},
  volume={5},
  pages={194},
  year={2022},
  doi={10.1038/s41746-022-00742-2}
}

@article{wiest2024,
  title={Privacy-Preserving Large Language Models for Structured Medical Information Retrieval},
  author={Wiest, Isabella Catharina and Ferber, Dyke and Zhu, Jiefu and van Treeck, Marko and Meyer, Sonja K. and Juglan, Radhika and Carrero, Zunamys I. and Paech, Daniel and Kleesiek, Jens and Ebert, Matthias P. and Truhn, Daniel and Kather, Jakob Nikolas},
  journal={npj Digital Medicine},
  volume={7},
  pages={257},
  year={2024},
  doi={10.1038/s41746-024-01233-2}
}

@article{wiest2025llmaix,
  title={A Software Pipeline for Medical Information Extraction with Large Language Models, Open Source and Suitable for Oncology},
  author={Wiest, Isabella Catharina and Wolf, Fabian and Lessmann, Marie-Elisabeth and van Treeck, Marko and Ferber, Dyke and Zhu, Jiefu and Boehme, Heiko and Bressem, Keno K. and Ulrich, Hannes and Ebert, Matthias P. and Kather, Jakob Nikolas},
  journal={npj Precision Oncology},
  volume={9},
  pages={313},
  year={2025},
  doi={10.1038/s41698-025-01103-4}
}

@article{griot2025,
  title={Implementation of Large Language Models in Electronic Health Records},
  author={Griot, Maxime and Vanderdonckt, Jean and Yuksel, Demet},
  journal={PLOS Digital Health},
  volume={4},
  number={12},
  pages={e0001141},
  year={2025},
  doi={10.1371/journal.pdig.0001141}
}

@inproceedings{lewis2020,
  title={Retrieval-Augmented Generation for Knowledge-Intensive {NLP} Tasks},
  author={Lewis, Patrick and Perez, Ethan and Piktus, Aleksandra and Petroni, Fabio and Karpukhin, Vladimir and Goyal, Naman and K{\"u}ttler, Heinrich and Lewis, Mike and Yih, Wen-tau and Rockt{\"a}schel, Tim and Riedel, Sebastian and Kiela, Douwe},
  booktitle={Advances in Neural Information Processing Systems},
  volume={33},
  pages={9459--9474},
  year={2020},
  doi={10.5555/3495724.3496517}
}

@inproceedings{yao2023,
  title={{ReAct}: Synergizing Reasoning and Acting in Language Models},
  author={Yao, Shunyu and Zhao, Jeffrey and Yu, Dian and Du, Nan and Shafran, Izhak and Narasimhan, Karthik and Cao, Yuan},
  booktitle={International Conference on Learning Representations},
  year={2023},
  doi={10.48550/arXiv.2210.03629}
}

@inproceedings{xiong2024,
  title={Improving Retrieval-Augmented Generation in Medicine with Iterative Follow-up Questions},
  author={Xiong, Guangzhi and Jin, Qiao and Wang, Xiao and Zhang, Minjia and Lu, Zhiyong and Zhang, Aidong},
  booktitle={Pacific Symposium on Biocomputing},
  volume={30},
  pages={199--214},
  year={2025},
  doi={10.1142/9789819807024\_0015}
}

@inproceedings{fayyaz2025,
  title={Collapse of Dense Retrievers: {Short}, Early, and Literal Biases Outranking Factual Evidence},
  author={Fayyaz, Mohsen and Modarressi, Ali and Schuetze, Hinrich and Peng, Nanyun},
  booktitle={Proceedings of the 63rd Annual Meeting of the Association for Computational Linguistics (Volume 1: Long Papers)},
  pages={9136--9152},
  year={2025},
  address={Vienna, Austria},
  publisher={Association for Computational Linguistics},
  doi={10.18653/v1/2025.acl-long.447}
}

@article{palm2025,
  title={Leveraging Interoperable Electronic Health Record ({EHR}) Data for Distributed Analyses in Clinical Research: Technical Implementation Report of the {HELP} Study},
  author={Palm, Julia and Saleh, Kutaiba and Scherag, Andr{\'e} and Ammon, Danny},
  journal={JMIR Medical Informatics},
  volume={13},
  number={1},
  pages={e68171},
  year={2025},
  doi={10.2196/68171}
}

@article{vorisek2022,
  title={Fast Healthcare Interoperability Resources ({FHIR}) for Interoperability in Health Research: Systematic Review},
  author={Vorisek, Carina Nina and Lehne, Moritz and Klopfenstein, Sophie Anne Ines and Mayer, Paula Josephine and Bartschke, Alexander and Haese, Thomas and Thun, Sylvia},
  journal={JMIR Medical Informatics},
  volume={10},
  number={7},
  pages={e35724},
  year={2022},
  doi={10.2196/35724}
}

@article{gruenig2025,
  title={Implementation and User Evaluation of an On-Premise Large Language Model in a {German} University Hospital Setting: Cross-Sectional Survey},
  author={Gr{\"u}nig, Ali{\'c}e and Kriebel, Jenifer and Varghese, Julian and Herrmann, Tim and Sandmann, Sarah and Bruns, Christian},
  journal={JMIR AI},
  volume={5},
  pages={e84362},
  year={2026},
  doi={10.2196/84362}
}

@misc{hl7fhir,
  author={{HL7}},
  title={Fast Healthcare Interoperability Resources ({FHIR}) Release 4},
  year={2019},
  howpublished={\url{https://hl7.org/fhir/R4/}},
  note={Accessed: 2026-06-15}
}

@article{pator2026,
  title={Serialisation Strategy Matters: How {FHIR} Data Format Affects {LLM} Medication Reconciliation},
  author={Pator, Sanjoy},
  journal={arXiv preprint arXiv:2604.21076},
  year={2026},
  doi={10.48550/arXiv.2604.21076}
}

@inproceedings{tam2024,
  title={Let Me Speak Freely? {A} Study on the Impact of Format Restrictions on Large Language Model Performance},
  author={Tam, Zhi Rui and Wu, Cheng-Kuang and Tsai, Yi-Lin and Lin, Chieh-Yen and Lee, Hung-yi and Chen, Yun-Nung},
  booktitle={Proceedings of the 2024 Conference on Empirical Methods in Natural Language Processing: Industry Track},
  pages={1218--1236},
  year={2024},
  address={Miami, Florida, US},
  publisher={Association for Computational Linguistics},
  doi={10.18653/v1/2024.emnlp-industry.91}
}

@article{clines2025,
  title={{CLINES}: Clinical {LLM}-based Information Extraction and Structuring Agent},
  author={Yang, Zongxin and Yuan, Hongyi and Sayeed, Raheel and Tan, Amelia Li Min and Cai, Enci and Moro, Mohammed and Li, Xiudi and Ying, Huaiyuan and Brown, Nicholas and Weber, Griffin and Yu, Sheng and Kohane, Isaac and Cai, Tianxi},
  journal={medRxiv},
  year={2025},
  doi={10.64898/2025.12.01.25341355},
  note={Preprint}
}

@article{harmone2025,
  title={{HARMON-E}: Hierarchical Agentic Reasoning for Multimodal Oncology Notes to Extract Structured Data},
  author={Gupta, Shashi Kant and Pramanik, Arijeet and Thomas, Jerrin John and Schwind, Regina and Wiener, Lauren and Raju, Avi and Kornbluth, Jeremy and Wang, Yanshan and Su, Zhaohui and Singh, Hrituraj},
  journal={arXiv preprint arXiv:2512.19864},
  year={2025},
  doi={10.48550/arXiv.2512.19864}
}

@inproceedings{reflectool2025,
  title={{ReflecTool}: Towards Reflection-Aware Tool-Augmented Clinical Agents},
  author={Liao, Yusheng and Jiang, Shuyang and Wang, Yanfeng and Wang, Yu},
  booktitle={Proceedings of the 63rd Annual Meeting of the Association for Computational Linguistics (Volume 1: Long Papers)},
  pages={13507--13531},
  year={2025},
  address={Vienna, Austria},
  publisher={Association for Computational Linguistics},
  doi={10.18653/v1/2025.acl-long.663}
}

@article{moon2022,
  title={Identifying Information Gaps in Electronic Health Records by Using Natural Language Processing: {Gynecologic} Surgery History Identification},
  author={Moon, Sungrim and Carlson, Luke A. and Moser, Ethan D. and Agnikula Kshatriya, Bhavani Singh and Smith, Carin Y. and Rocca, Walter A. and Gazzuola Rocca, Liliana and Bielinski, Suzette J. and Liu, Hongfang and Larson, Nicholas B.},
  journal={Journal of Medical Internet Research},
  volume={24},
  number={1},
  pages={e29015},
  year={2022},
  doi={10.2196/29015}
}

\appendix

\section{Clinical Study Extraction Schema}
\label{sec:appendix-study}

Table~\ref{tab:study} lists the 74 AI-extracted fields of the lymphoma registry study, grouped by data type; the type determines how the agent is prompted and how its output is validated. Four further demographic fields are read directly from FHIR and are excluded from the evaluation in \S\ref{sec:study-eval}. The eCRF was designed by two nuclear-medicine physicians and a lymphoma hematologist, and it spans the full disease trajectory, from diagnosis and histopathology through molecular characterization, treatment, imaging follow-up, and outcomes. Related single-marker fields are grouped into one row where they share a common form; every field name is listed explicitly.

\begin{table*}
\centering
\small
\begin{tabular}{p{4.7cm}p{10.3cm}}
\toprule
\textbf{Type ($n$)} & \textbf{Fields} \\
\midrule
\textbf{Categorical (45)} & \\
\quad WHO classification \& subtyping (9) & Primary lymphoma category; entity-specific subtype for LBCL, DLBCL cell-of-origin, follicular, Hodgkin, mantle-cell, marginal-zone, Burkitt, and peripheral T-/NK-cell lymphoma \\
\quad Histology \& qualitative IHC (9) & Biopsy sites; necrosis; positive/negative immunohistochemistry for p53, PD-L1 (tumour cells), MYC, BCL2, CD20, CD30, and EBV (EBER ISH) \\
\quad Cytogenetics \& FISH (9) & Cytogenetics performed; karyotype available; complex karyotype; FISH status; MYC / BCL2 / BCL6 rearrangement; del(17p)/TP53; 9p24.1 amplification \\
\quad NGS mutation status (17) & Overall NGS status; TP53, MYD88 (L265P), NOTCH1, NOTCH2, EZH2, CD79B, ARID1A, MEF2B, EP300, FOXO1, CREBBP, CARD11, RHOA (G17V), TET2, DNMT3A, IDH2 (R172) \\
\quad Outcome (1) & Overall survival event \\
\midrule
\textbf{Numerical (9)} & Height; weight at diagnosis; months from ASCT to relapse or progression; Ki-67 index; IHC expression percentage for p53, PD-L1 (tumour cells), PD-L1 (immune cells), MYC, and BCL2 \\
\midrule
\textbf{Boolean (8)} & Significant comorbidities (CIRS); B-symptoms; progression event; relapse event; prior ASCT before relapse/progression; histological transformation; bone-marrow involvement; large-cell component \\
\midrule
\textbf{Date (6)} & Diagnosis; most recent biopsy; transformation; progression or last follow-up; relapse or last follow-up; death or last follow-up \\
\midrule
\textbf{Free text (3)} & Original pathology-report diagnosis; extranodal site specification; biopsied body-fluid specification \\
\midrule
\textbf{Tabular (3)} & Treatment timeline (each therapy line, transplant, surgery, and radiation with dates and cycles); PET examinations (dates, indication, tracer); MRD assessments (method, result) \\
\bottomrule
\end{tabular}
\caption{The 74 AI-extracted fields of the lymphoma study schema, grouped by data type. Parenthesized counts give the number of fields. Related single-marker categorical fields are summarized by group; every field name is listed.}
\label{tab:study}
\end{table*}

\section{Rejection Categories}
\label{sec:appendix-rejcat}

When a reviewer rejects an extracted value, they assign a category describing the nature of the problem. The categories fall into three groups, mirroring Table~\ref{tab:rejections}.

\textbf{Extraction errors} mean the value is factually wrong, unsupported, or absent. \emph{Incorrect value}: the value must be partially corrected. \emph{Fully incorrect}: the value is wrong and must be replaced wholesale, though it remains attributed to a cited passage, which distinguishes it from a hallucination. \emph{Outdated}: a once-correct value that is no longer current. \emph{Missed extraction}: the system returned nothing although the information is present in the sources. \emph{Extraneous extraction}: the converse, a value was produced for a field that should have been left empty. \emph{Hallucinated}: content not supported by the cited source. \emph{Missing reference}: the value or a crucial detail is given without attribution to a source passage.

\textbf{Editorial adjustments} mean the value is acceptable but the amount or form of information differs from what the field wanted. \emph{Missing information}: available, relevant detail was omitted. \emph{Excess information}: more was returned than the field asked for. \emph{Reformatting}: the value is correct but formatted incorrectly.

\textbf{Form configuration} covers rejections that reflect the form rather than the extraction. \emph{Configuration error}: the field's options or definition did not match clinical reality, a form-design issue rather than an extraction failure.

\section{Error Analysis}
\label{sec:appendix-erroranalysis}

This appendix breaks the 253 rejections down three ways. Table~\ref{tab:abstention} separates each field type's acceptance into the case where the system returned a value (precision) and the case where it abstained (abstention reliability). Table~\ref{tab:rejbytype} locates each rejection category within the field types. Table~\ref{tab:topfields} shows how concentrated the errors are: ten of the 74 fields account for 182 of the 253 rejections.

\begin{table}
\centering
\small
\begin{tabular}{lrrrr}
\toprule
& \multicolumn{2}{c}{\textbf{Returned a value}} & \multicolumn{2}{c}{\textbf{Returned empty}} \\
\cmidrule(lr){2-3}\cmidrule(lr){4-5}
\textbf{Field type} & $n$ & Acc.\% & $n$ & Acc.\% \\
\midrule
Categorical & 2,892 & 98.3 & 1,563 & 99.1 \\
Numerical & 213 & 98.1 & 678 & 98.4 \\
Boolean & 619 & 98.4 & 173 & 99.4 \\
Free text & 119 & 93.3 & 178 & 97.8 \\
Date & 392 & 91.8 & 202 & 69.8 \\
Tabular & 205 & 71.2 & 92 & 98.9 \\
\midrule
\textbf{Total} & \textbf{4,440} & \textbf{96.4} & \textbf{2,886} & \textbf{96.8} \\
\bottomrule
\end{tabular}
\caption{Acceptance split by whether the system returned a value or abstained. Tabular fields fail when they produce a value (71.2\%) but abstain reliably; dates are the reverse, abstaining unreliably (69.8\%). All other types are strong in both modes.}
\label{tab:abstention}
\end{table}

\begin{table*}
\centering
\small
\begin{tabular}{lrrrrrrrr r}
\toprule
\textbf{Field type} & \textbf{Rej.} & \textbf{Incorrect} & \textbf{Fully incorr.} & \textbf{Miss.\ info} & \textbf{Excess} & \textbf{Missed} & \textbf{Extran.} & \textbf{Miss.\ ref.} & \textbf{Config.} \\
\midrule
Categorical & 62 & 48 & 1 & 1 & -- & 3 & 1 & 1 & 7 \\
Numerical & 15 & 12 & -- & -- & -- & 2 & -- & 1 & -- \\
Boolean & 11 & 6 & 2 & 1 & -- & -- & -- & -- & 2 \\
Date & 93 & 89 & 2 & -- & -- & 1 & -- & 1 & -- \\
Free text & 12 & 11 & -- & -- & -- & 1 & -- & -- & -- \\
Tabular & 60 & 55 & 2 & -- & 1 & 2 & -- & -- & -- \\
\midrule
\textbf{Total} & \textbf{253} & \textbf{221} & \textbf{7} & \textbf{2} & \textbf{1} & \textbf{9} & \textbf{1} & \textbf{3} & \textbf{9} \\
\bottomrule
\end{tabular}
\caption{Rejection category by field type (counts). Categories not triggered anywhere (hallucinated, outdated, reformatting) are omitted; definitions in Appendix~\ref{sec:appendix-rejcat}. Configuration errors occur only in categorical and Boolean fields (form-option mismatches); dates are almost entirely incorrect-value.}
\label{tab:rejbytype}
\end{table*}

\begin{table}
\centering
\small
\begin{tabular}{p{4.6cm}lr}
\toprule
\textbf{Field} & \textbf{Type} & \textbf{Rej.} \\
\midrule
Date of death / last follow-up & Date & 44 \\
Treatment timeline & Tabular & 40 \\
Date of progression / last PET & Date & 22 \\
Overall survival event & Categorical & 16 \\
MRD assessments & Tabular & 15 \\
Weight at diagnosis & Numerical & 12 \\
Date of relapse / last PET & Date & 11 \\
Biopsy sites & Categorical & 8 \\
Diagnosis date & Date & 7 \\
Extranodal site specification & Free text & 7 \\
\bottomrule
\end{tabular}
\caption{The ten most-rejected fields (each evaluated on all 99 patients), accounting for 182 of the 253 rejections. Date and table fields dominate the error mass.}
\label{tab:topfields}
\end{table}

\section{Per-Patient Context Distributions}
\label{sec:appendix-perpatient}

Table~\ref{tab:perpatient-full} provides the full distribution of per-patient context statistics. The mean consistently exceeds the median across all dimensions, reflecting heavy right skew: a minority of patients accumulate disproportionately large contexts. The P1/P99 columns bound the central 98\% of patients. The gap from P99 to Max defines the hardest cases the agentic framework must handle. The top 1\% of patients hold at least 937 documents and 37,074 structured resources, reaching up to 2,542 and 119,191 respectively. The history length maximum (739,726 days) reflects corrupt timestamps in the source data. P99 (9,775 days, $\sim$26.8 years) provides a more realistic upper bound.

\begin{table*}
\centering
\small
\begin{tabular}{lrrrrrrrr}
\toprule
& \textbf{Min} & \textbf{P1} & \textbf{Q1} & \textbf{Med.} & \textbf{Q3} & \textbf{P99} & \textbf{Max} & \textbf{Mean} \\
\midrule
Raw docs & 1 & 2 & 22 & 77 & 193 & 1,269 & 3,029 & 164.6 \\
Deduped & 1 & 1 & 14 & 52 & 140 & 937 & 2,542 & 120.4 \\
Encounters & 0 & 3 & 6 & 18 & 46 & 323 & 1,207 & 42.0 \\
Structured & 0 & 1 & 38 & 406 & 1,922 & 37,074 & 119,191 & 2,804.1 \\
History (days) & 0 & 15 & 330 & 1,305 & 4,145 & 9,775 & 739,726 & 2,699.7 \\
\midrule
Doc.\ length (chars)$^\dagger$ & 24 & -- & 1,387 & 2,915 & 11,035 & -- & 907,179 & 10,788 \\
\bottomrule
\end{tabular}
\caption{Per-patient context statistics ($n$=10,000). ``Structured'' counts all non-document FHIR resources (lab values, medications, conditions, etc.). Table~\ref{tab:patients} in the main text shows the compressed version. $^\dagger$The last row is per document, not per patient, over the OCR-processed documents (P1/P99 unavailable); document length spans four orders of magnitude, so a fixed chunking or truncation budget cannot serve both ends.}
\label{tab:perpatient-full}
\end{table*}

\section{Encounter Coverage by Patient Complexity}
\label{sec:appendix-enc-coverage}

Table~\ref{tab:enc-coverage-bucket} shows how top-encounter coverage varies with the number of encounters per patient. For patients with few encounters ($\leq$5), a single encounter typically holds all documents. As encounters increase, coverage disperses: for patients with 20+ case-level encounters, the median drops to 14.7\%, suggesting the hierarchy distributes documents across episodes as intended. Yet at P99, a single encounter still holds 53.5\% of documents. The concentration index (Table~\ref{tab:concentration}) confirms this pattern, reaching 14.83 at P99: the patients with the most complex records are precisely those where documents cluster most unevenly. Encounter-based scoping would therefore degrade for exactly the patients that depend most on thorough retrieval.

\begin{table*}
\centering
\small
\begin{tabular}{lrrrrrrrrr}
\toprule
\textbf{Case enc.} & \textbf{$n$} & \textbf{Min} & \textbf{P1} & \textbf{Q1} & \textbf{Median} & \textbf{Q3} & \textbf{P99} & \textbf{Max} & \textbf{Mean} \\
\midrule
1 & 1,930 & 9.5 & 42.9 & 100.0 & 100.0 & 100.0 & 100.0 & 100.0 & 96.6 \\
2--5 & 3,334 & 2.1 & 23.8 & 46.7 & 59.0 & 75.0 & 100.0 & 100.0 & 61.3 \\
6--20 & 3,099 & 6.9 & 12.0 & 24.3 & 32.6 & 44.0 & 80.2 & 100.0 & 35.5 \\
20+ & 1,637 & 2.5 & 3.9 & 10.0 & 14.7 & 21.1 & 53.5 & 85.5 & 17.1 \\
\bottomrule
\end{tabular}
\caption{Top-encounter document coverage (\%) by number of case-level encounters per patient. Shows the percentage of a patient's documents held by their single busiest encounter. Even for patients with 20+ encounters, P99 coverage remains 53.5\%, indicating persistent clustering.}
\label{tab:enc-coverage-bucket}
\end{table*}

Table~\ref{tab:concentration} gives the full distribution of top-encounter coverage and of the concentration index used in \S\ref{sec:dataquality}.

\begin{table*}
\centering
\small
\begin{tabular}{lrrrrrrrr}
\toprule
& \textbf{Min} & \textbf{P1} & \textbf{Q1} & \textbf{Med.} & \textbf{Q3} & \textbf{P99} & \textbf{Max} & \textbf{Mean} \\
\midrule
Top-encounter coverage (\%) & 2.1 & 6.3 & 26.7 & 47.5 & 80.0 & 100.0 & 100.0 & 52.9 \\
Concentration index & 0.04 & 0.73 & 1.27 & 2.24 & 3.75 & 14.83 & 105.1 & 3.07 \\
\bottomrule
\end{tabular}
\caption{Per-patient distribution of top-encounter document coverage ($n$=10,000) and the concentration index ($n$=9,957 patients with at least one case-level encounter). The concentration index is the ratio of the top encounter's document share to the uniform expectation across case-level encounters; 1.0 indicates even distribution.}
\label{tab:concentration}
\end{table*}

\section{Patient History Length}
\label{sec:appendix-history}

Tables~\ref{tab:history} and~\ref{tab:resources-history} break down document counts and total FHIR resources by patient history length. Both show heavy right skew, with the gap from P99 to Max again highlighting the extreme cases the system must handle.

\begin{table*}
\centering
\small
\begin{tabular}{lrrrrrrrrr}
\toprule
\textbf{History} & \textbf{$n$} & \textbf{Min} & \textbf{P1} & \textbf{Q1} & \textbf{Med.} & \textbf{Q3} & \textbf{P99} & \textbf{Max} & \textbf{Mean} \\
\midrule
$<$ 1 yr & 2,709 & 1 & 1 & 4 & 10 & 26 & 190 & 497 & 23.7 \\
1--3 yr & 1,995 & 1 & 2 & 18 & 49 & 102 & 572 & 1,476 & 86.0 \\
3--5 yr & 934 & 2 & 6 & 34 & 82 & 162 & 777 & 1,360 & 132.6 \\
5--10 yr & 1,602 & 1 & 6 & 50 & 107 & 235 & 1,217 & 2,542 & 187.9 \\
10+ yr & 2,760 & 1 & 3 & 43 & 114 & 252 & 1,174 & 2,439 & 196.8 \\
\bottomrule
\end{tabular}
\caption{Deduplicated documents per patient by history length ($n$=10,000).}
\label{tab:history}
\end{table*}

\begin{table*}
\centering
\small
\begin{tabular}{lrrrrrrrrr}
\toprule
\textbf{History} & \textbf{$n$} & \textbf{Min} & \textbf{P1} & \textbf{Q1} & \textbf{Med.} & \textbf{Q3} & \textbf{P99} & \textbf{Max} & \textbf{Mean} \\
\midrule
$<$ 1 yr & 2,709 & 1 & 5 & 17 & 52 & 291 & 13,040 & 52,462 & 711.4 \\
1--3 yr & 1,995 & 6 & 8 & 105 & 490 & 1,645 & 29,889 & 102,412 & 2,241.7 \\
3--5 yr & 934 & 12 & 20 & 225 & 931 & 2,820 & 30,848 & 83,760 & 2,996.0 \\
5--10 yr & 1,602 & 5 & 19 & 352 & 1,177 & 4,061 & 47,994 & 100,673 & 4,288.7 \\
10+ yr & 2,760 & 6 & 18 & 365 & 1,303 & 4,939 & 48,929 & 121,542 & 5,086.6 \\
\bottomrule
\end{tabular}
\caption{Total FHIR resources per patient by history length ($n$=10,000). Includes all non-document resources (lab values, medications, conditions, encounters, etc.).}
\label{tab:resources-history}
\end{table*}

\section{Metadata and Timestamp Population}
\label{sec:appendix-metadata}

Tables~\ref{tab:metadata-population} and~\ref{tab:timestamp-population} report the population rates of document-level metadata and timestamp fields across the two FHIR resource types that carry clinical documents. These rates quantify the metadata sparsity discussed in \S\ref{sec:dataquality}.

\begin{table}
\centering
\small
\begin{tabular}{llr}
\toprule
\textbf{Resource} & \textbf{Metadata Field} & \textbf{Pop. \%} \\
\midrule
\multicolumn{3}{l}{\emph{Identification and relationships}} \\
DocRef & Unique identifier & 27.8 \\
DocRef & Related documents & 0.52 \\
\midrule
\multicolumn{3}{l}{\emph{Authorship and provenance}} \\
DocRef & Author & 1.9 \\
DocRef & Authenticator & 16.2 \\
DocRef & Custodian & 100.0 \\
DiagRep & Performer & 27.1 \\
DiagRep & Results interpreter & 97.5 \\
\midrule
\multicolumn{3}{l}{\emph{Content description}} \\
DocRef & Description & 99.7 \\
DocRef & Attachment title & 0.0 \\
DocRef & Content format & 98.1 \\
DiagRep & Title & 60.6 \\
DiagRep & Structured conclusion & 0.45 \\
\bottomrule
\end{tabular}
\caption{Document metadata population rates. DocRef: DocumentReference ($n$=636,534); DiagRep: DiagnosticReport ($n$=567,110). ``Pop.~\%'' is the fraction of resources where the field is non-empty.}
\label{tab:metadata-population}
\end{table}

\begin{table}
\centering
\small
\begin{tabular}{llr}
\toprule
\textbf{Resource} & \textbf{Timestamp Field} & \textbf{Pop. \%} \\
\midrule
DocRef & Report date & 99.99 \\
DocRef & File creation date & 100.0 \\
DocRef & Encounter period & 0.0 \\
DocRef & Release date & 76.5 \\
DocRef & Print date & 24.7 \\
DocRef & Record last updated & 100.0 \\
\midrule
DiagRep & Effective date & 95.2 \\
DiagRep & Issued date & 78.0 \\
DiagRep & File creation date & 97.7 \\
DiagRep & Record last updated & 100.0 \\
\bottomrule
\end{tabular}
\caption{Timestamp field population rates. DocRef: DocumentReference ($n$=636,534); DiagRep: DiagnosticReport ($n$=567,110). ``Pop.~\%'' is the fraction of resources where the field is non-empty.}
\label{tab:timestamp-population}
\end{table}

\section{Timestamp Cross-Validation}
\label{sec:appendix-crossval}

Table~\ref{tab:crossval} compares the FHIR metadata timestamp resolved for each document against the clinical date extracted from the document content via OCR and an LLM ($n$=15,142 documents). The alignment is consistent at roughly 59\% regardless of which FHIR field provides the resolved date, confirming that no single metadata timestamp reliably represents when clinical activity occurred.

\begin{table}
\centering
\small
\begin{tabular}{lrrrr}
\toprule
\textbf{FHIR Field} & \textbf{$n$} & \textbf{Same day} & \textbf{$\pm$1 day} & \textbf{$>$1 day} \\
\midrule
All fields & 15,142 & 58.8\% & 4.8\% & 36.5\% \\
\midrule
Report date & 8,037 & 58.6\% & 5.2\% & 36.2\% \\
Effective date & 6,240 & 59.0\% & 3.7\% & 37.3\% \\
Issued date & 859 & 59.1\% & 7.9\% & 32.9\% \\
File creation & 6 & 50.0\% & 0.0\% & 50.0\% \\
\bottomrule
\end{tabular}
\caption{Alignment between FHIR metadata timestamps and clinical dates extracted from document content. Rows show the FHIR field that provided the resolved date for each document. Overall same-day agreement: 58.8\%.}
\label{tab:crossval}
\end{table}

\end{document}